%% file: main.tex
\documentclass[pdflatex,sn-basic,twocolumn]{sn-jnl}

\usepackage{graphicx}%
\usepackage{multirow}%
\usepackage{amsmath,amssymb,amsfonts}%
\usepackage{mathrsfs}%
\usepackage[title]{appendix}%
\usepackage[table]{xcolor}%
\usepackage{textcomp}%
\usepackage{manyfoot}%
\usepackage{booktabs}%
\usepackage{algorithm}%
\usepackage{algorithmicx}%
\usepackage{algpseudocode}%
\usepackage{listings}%
\usepackage{algcompatible}
\usepackage{makecell}
\usepackage{arydshln}
\usepackage{subcaption}
\usepackage{colortbl}
\usepackage{hyperref}
\usepackage{threeparttable}
\input{definitions}
\newgeometry{top=1in, bottom=1.5in, left=1in, right=1in, twocolumn}

\usepackage{anyfontsize}
\usepackage{longtable}
\usepackage[sorting=none, maxnames=5, minnames=1, backend=bibtex, maxcitenames=1,mincitenames=1]{biblatex}
\newcommand{\citet}{\textcite}
\usepackage{pdfpages} 
\usepackage{pgffor} 
\newif\ifarXiv
\arXivtrue 
\def\supplementfilename{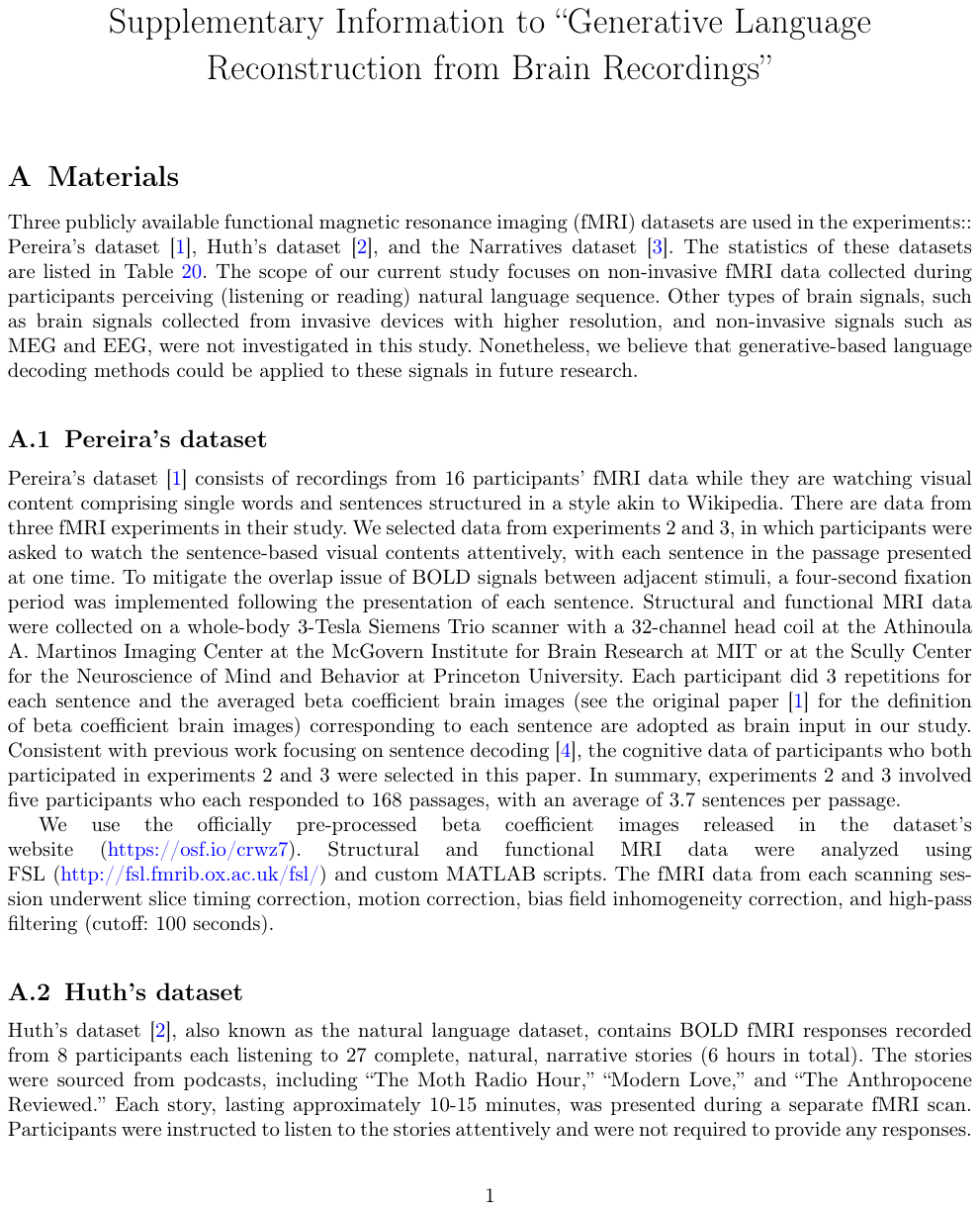}
\def\numbersupplementpages{64}

\begin{document}

\title[Article Title]{BrainLLM: Generative Language Decoding from Brain Recordings}

\author[1]{\fnm{Ziyi} \sur{Ye}}\email{yeziyi1998@gmail.com}
\author[1]{\fnm{Qingyao} \sur{Ai}}\email{aiqy@tsinghua.edu.cn}
\author*[1]{\fnm{Yiqun} \sur{Liu}}\email{yiqunliu@tsinghua.edu.cn}
\author[2]{\fnm{Maarten} \sur{de Rijke}}\email{M.deRijke@uva.nl}
\author[1]{\fnm{Min} \sur{Zhang}}\email{z-m@tsinghua.edu.cn}
\author[3]{\fnm{Christina} \sur{Lioma}}\email{c.lioma@di.ku.dk}
\author[3,4]{\fnm{Tuukka} \sur{Ruotsalo}}\email{tr@di.ku.dk}

\affil*[1]{\orgname{Tsinghua University}, Beijing, China}
\affil[2]{\orgname{University of Amsterdam}, Amsterdam, Netherlands}
\affil[3]{\orgname{University of Copenhagen}, Copenhagen, Denmark}
\affil[4]{\orgname{Lappeenranta-Lahti University of Technology}, Lappeenranta, Finland}

\abstract{
\input{meta/abstract}
}


\maketitle

\input{meta/1_introduction}
\input{meta/3_results}
\input{meta/3_2_subresults}

\input{meta/4_discussion}
\input{meta/5_method}

\subsection*{Acknowledgements}
This work is supported by Quan Cheng Laboratory~(Grant No. QCLZD202301), the Academy of Finland, the Horizon 2020 FET program of the EU through the ERA-NET Cofund funding grant CHIST-ERA-20-BCI-001, and the University of Copenhagen.
The authors sincerely acknowledge the Members of the IRLab at the University of Copenhagen and Tsinghua University for their comments and help.
Additionally, we appreciate the manuscript reviewers for their constructive suggestions and feedback. 

\subsection*{Author contributions}
ZY contributed conceptualization, methodology, experiments, and writing. TR, CL, \& QA contributed conceptualization, formal analysis, supervision, and writing. YL \& MZ contributed funding acquisition, resources, and supervision. MdR contributed formal analysis and writing.

\subsection*{Competing interests}
The authors declare no competing interests.

\clearpage
\printbibliography
\clearpage

\ifarXiv
    \foreach \x in {1,...,\numbersupplementpages}
    {
        \includepdf[pages={\x}]{\supplementfilename}
    }
\fi

\end{document}

%% file: definitions.tex
\usepackage{acronym}

\acrodef{MLP}{multilayer perceptron}
\acrodef{fMRI}{functional magnetic resonance imaging}
\acrodef{BCI}{brain–computer interface}
\acrodef{LLM}{large language model}
\acrodef{FDR}{false discovery rate}
\acrodef{NLP}{natural language processing}
\acrodef{TR}{time repetition}
\acrodef{MSE}{mean square error}
\acrodef{BOLD}{blood oxygen level dependent}

\definecolor{lightblue}{RGB}{173,216,230}
\definecolor{pere1}{HTML}{eaafaf}
\definecolor{huth1}{HTML}{d1c4a3}
\definecolor{nara1}{HTML}{8DB3BF}
\definecolor{pere2}{HTML}{f5d9d9}
\definecolor{huth2}{HTML}{eaeaea}
\definecolor{nara2}{HTML}{a8d5e2}

%% file: meta/abstract.tex
Language reconstruction from non-invasive brain recordings has been a long-standing challenge.
Existing research has addressed this challenge with a classification setup, where a set of language candidates are pre-constructed and then matched with the representation decoded from brain recordings.
Here, we propose a method that addresses language reconstruction through auto-regressive generation, which directly uses the representation decoded from functional magnetic resonance imaging~(fMRI) as the input for a large language model~(LLM), mitigating the need for pre-constructed candidates.
While an LLM can already generate high-quality content, our approach produces results more closely aligned with the visual or auditory language stimuli in response to which brain recordings are sampled, especially for content deemed ``surprising'' for the LLM. 
Furthermore, we show that the proposed approach can be used in an auto-regressive manner to reconstruct a 10-minute-long language stimulus.
Our method outperforms or is comparable to previous classification-based methods under different task settings, with the added benefit of estimating the likelihood of generating any semantic content.
Our findings demonstrate the effectiveness of employing brain language interfaces in a generative setup and delineate a powerful and efficient means for mapping functional representations of language perception in the brain.

%% file: meta/1_introduction.tex
\section*{Introduction}
Reconstruction of natural language from brain recordings not only provides potential insights into understanding how the human brain forms language, but also facilitates the development of neural communication interfaces for restorative and augmentative applications.
Previous work has demonstrated that it is possible to decode meaningful linguistic and semantic information from brain recordings to guide classification tasks, such as selecting a target from a set of words \cite{mitchell2008predicting,pei2011decoding}, sentences~\cite{pereira2018toward,tang2023semantic}, and topics~\cite{kivisaari2019reconstructing}. 
For instance, \citet{moses2021neuroprosthesis} successfully decoded the target words from a vocabulary of 50 words, using the brain recordings of an anarthria patient with electrodes implanted in the sensorimotor cortex. 
\citet{pereira2018toward} utilized non-invasive \ac{fMRI} data to decode the target sentence from a pair of or a set of sentences that were presented as visual stimuli.

Recently, \acp{LLM}, particularly those based on generative settings~\cite{radford2019language,brown2020language,touvron2023llama}, have become a dominant approach in computational language modeling. 
Those \acp{LLM} treat the process of language construction as a generation problem.
Given a text prompt, LLMs generate the most likely continuation based on the statistical semantic knowledge they learned from a vast amount of text.
By solving the language generation problem in an auto-regressive manner, \acp{LLM} can construct continuous texts that maintain both semantic and syntactic coherence \cite{touvron2023llama}. 
Leveraging the powerful capabilities of LLMs, recent language \acp{BCI}~\cite{tang2023semantic,affolter2020brain2word} have attempted to link LLMs with the decoding of brain signals.
For example, \citet{tang2023semantic} use an LLM to pre-construct a set of possible language candidates and then select the best one based on their similarities with the semantic representations decoded from the \ac{fMRI} data.

However, the methods listed above consider brain decoding and language generation as two separate phases.
Semantic representations extracted from brain recordings are used exclusively in a post-hoc selection phase.
While LLMs represent a leap forward in mimicking human language, they merely generate the most likely continuations based on their training material~\cite{radford2019language,brown2020language}.
In other words, there is no guarantee that the language generated by LLMs reflects the semantics decoded from brain recordings. 
Therefore, integrating brain recordings directly into the language generation process remains an open and unsolved challenge.
At the same time, a growing body of research highlights similarities between the representations and computational principles of language models and the human brain \cite{toneva2019interpreting,antonello2024predictive,goldstein2022shared}.
This suggests the potential to leverage brain representations as inputs to large language models.

\begin{figure*}[t]
    \centering
    \includegraphics[width=\textwidth]{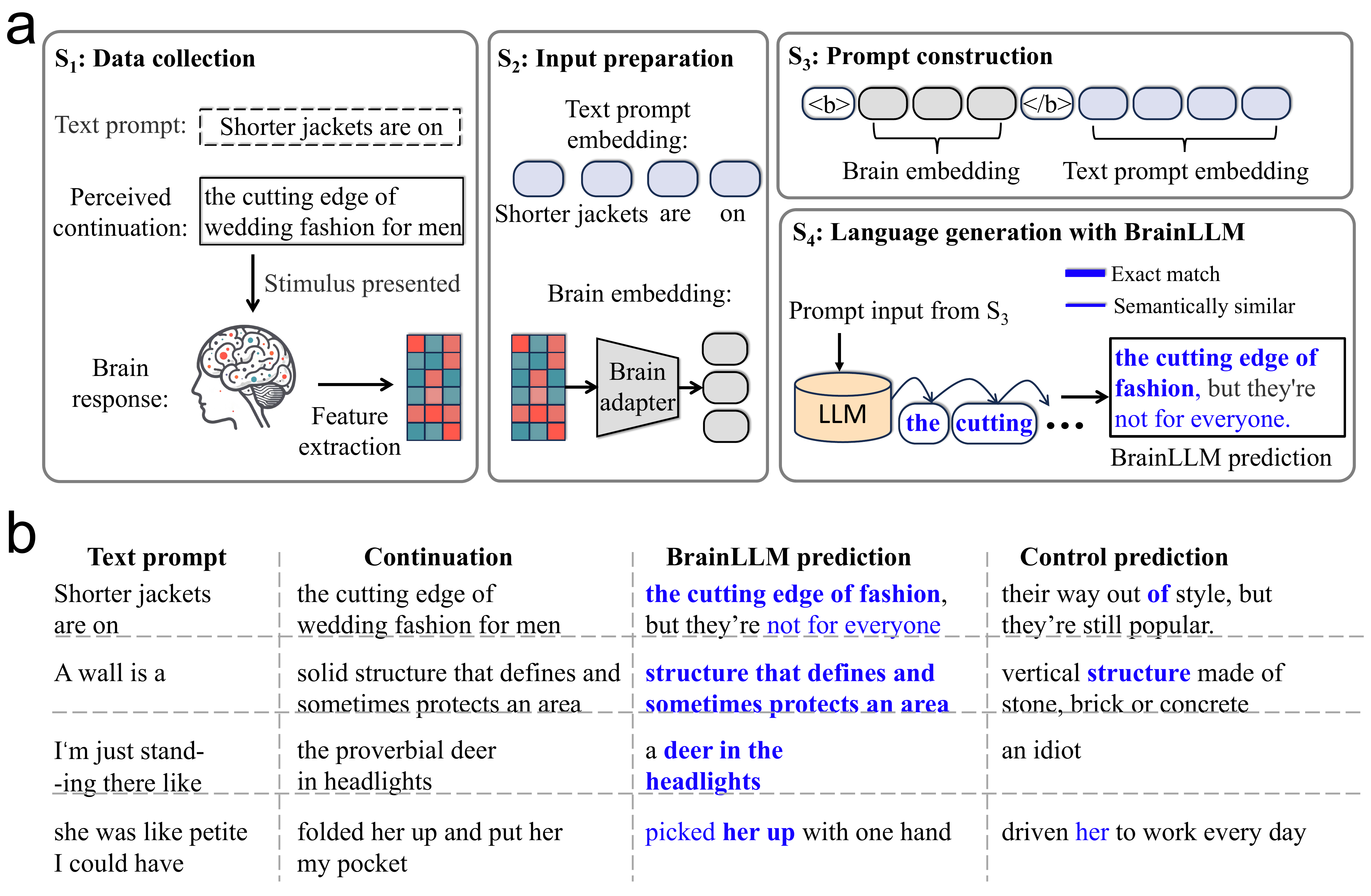}
    \caption{\textbf{Language generation with brain recordings~(BrainLLM).} \textbf{(a)} The generation process has four main stages. \textbf{$S_1$}: Brain recordings in response to the perceived continuation are collected. 
        \textbf{$S_2$}: A brain adapter extracts features from brain recordings and transforms them into hidden vectors that match the shape of text embeddings in a standard LLM. 
        \textbf{$S_3$}: Brain embeddings and text prompt embeddings are concatenated as a prompt input. 
        \textbf{$S_4$}: The prompt input is fed into the LLM for language generation.
        BrainLLM generates content that is an exact match~(``the cutting edge of'') with, or semantically similar/gist match content~(``not for everyone'') to the perceived continuation. 
        \textbf{(b)} Examples of language generation with BrainLLM and its controls~(PerBrainLLM). 
        Text in \textcolor{blue}{blue} and \textbf{\textcolor{blue}{bold}} indicates that the generated content and the ground truth~(perceived continuation) are manually annotated as semantically similar and an exact match, respectively. }
    \label{fig:model_structure2}
\end{figure*}

Here, we present BrainLLM, an approach in which the semantic representation decoded from brain recordings is directly involved in the generation phase of continuous language. 
We focus on language generation from non-invasive \ac{fMRI} recordings of healthy participants perceiving visual or auditory language stimuli.
As depicted in Fig.~\ref{fig:model_structure2}, our proposed model generates a continuation of language from a given text prompt~(See Supplementary Tables~1--9, and Supplementary Figs.~1-3 for additional examples).
Unlike existing work~\cite{tang2023semantic,affolter2020brain2word}, BrainLLM incorporates brain signals directly in the language generation phase, thereby eliminating the need for post-hoc selection among pre-constructed language candidates. 
This approach significantly improves performance over standard LLM generation with only the text prompt, and over methods that use a pre-construction and post-hoc selection setup.
In addition, this method provides potential applications for neuroscience and machine learning research.
For example, BrainLLM can facilitate the investigation of linguistic encoding in the human brain by accessing the generation likelihood of any language content with various characteristics instead of a limited number of pre-defined candidates.

To accomplish this, BrainLLM consists of four key steps illustrated in Fig.~\ref{fig:model_structure2}(a): 
(1)~brain data is collected and features are extracted;
(2)~a brain adapter learns an embedding from the brain recordings;
(3)~prompts are constructed from brain and text modalities; 
(4)~language is generated in an auto-regressive manner based on a model of the prompt and an \ac{LLM}.
The brain adapter learns to map the space of brain representations onto a space with the same dimensionality as the text embeddings in the LLM.
This facilitates the generation based on a prompt representation that integrates both the brain modality and the text modality.
A protocol called ``prompt tuning''~\cite{liu2023gpt} and a generation-based loss function is adopted to train the brain adapter.
This protocol guarantees that the parameters in the \acp{LLM} are fixed while only the brain adapter is updated during training. 
To this end, the model parameters of the decoder can be fully trained with only a limited amount of neurological data compared to the data size typically used for training an \ac{LLM}.

%% file: meta/3_results.tex
\section*{Results}
\label{sec:results}
We evaluate BrainLLM using three \ac{fMRI} datasets~\cite{pereira2018toward,nastase2021narratives,lebel2023natural} in which participants perceive visual or auditory language stimuli~(see Supplementary Information A).
We construct a language generation task for each time frame~(e.g., a \ac{TR} of 2s in Huth's dataset) during the \ac{fMRI} recording process. 
As depicted in Fig.~\ref{fig:model_structure2}, the preceding text~(if any) to a time frame serves as the text prompt~(see Method). 
Meanwhile, the presented language stimulus within the time frame is considered as the perceived continuation, typically encompassing 3--10 words.
Then, the model's generation ability is evaluated by aligning its generation output to the perceived continuation.
We trained and evaluated the model for each human participant, involving 5 participants in Pereira's dataset~\cite{pereira2018toward}, 8 participants in Huth's dataset~\cite{lebel2023natural}, and 28 participants in the Narratives dataset~\cite{nastase2021narratives}.
We test BrainLLM's ability with the backbone LLM selected as Llama-2~\cite{touvron2023llama} because it is one of the best-performing public-sourced models.
Additionally, we extend our analysis to include the GPT-2 series~\cite{radford2019language} with varying sizes. 
A split-by-stimuli protocol is applied~(see Supplementary Information B.1) to ensure that the language stimuli and the corresponding brain response used during testing have not been seen in the training set.

\begin{table*}
    \centering
    \caption{
    \textbf{Language generation performance averaged across participants in different datasets.} }
    \label{tab:ml}
    \begin{tabular}{llllll}
    \toprule
     {\textbf{Dataset}}  &  \textbf{Model}&  \textbf{BLEU-1}($\uparrow$) &  \textbf{ROUGE-1}($\uparrow$) &  \textbf{ROUGE-L}($\uparrow$) & { \textbf{WER}($\downarrow$)}    \\
    \midrule
    \multirow{2}{*}{ Huth's}      
    & PerBrainLLM  & $\text{0.1668}^{*}$ & $\text{0.1536}^{*}$ & $\text{0.1474}^{*}$  & $\text{0.9109}^{*}$ \\
    & BrainLLM   & \textbf{0.1899} & \textbf{0.1780} & \textbf{0.1709}  & \textbf{0.8916} \\
    \midrule
    \multirow{2}{*}{Pereira's}   
    & {PerBrainLLM}  & $\text{0.3269}^{*}$ & $\text{0.2815}^{*}$ & $\text{0.2751}^{*}$  & $\text{0.7783}^{*}$ \\
    & BrainLLM & \textbf{0.3432} & \textbf{0.2987} & \textbf{0.2878}  & \textbf{0.7576} \\
    \midrule
    \multirow{2}{*}{Narratives} 
    & PerBrainLLM  & $\text{0.1269}^{*}$ &  $\text{0.1211}^{*}$ & $\text{0.1105}^{*}$  & $0.9311^{*}$ \\
    & BrainLLM   & \textbf{0.1375} & \textbf{0.1301} & \textbf{0.1209}  & \textbf{0.9239} \\
    \bottomrule
    \end{tabular}
    \begin{tablenotes}
        \item $*$ indicates that the difference between BrainLLM and PerBrainLLM is significant at $\ac{FDR} < 0.05$~(one-sided non-parametric test).
    \end{tablenotes}
\end{table*}

We conduct three evaluations to study the performance of BrainLLM: 
First, we compare the language reconstruction performance of BrainLLM to a control model PerBrainLLM, which randomly assigns the brain recordings as inputs across different prediction tasks through a permutation, and breaks connections between the stimuli and brain responses~(see Table~\ref{tab:ml} and Supplementary Information B.2).
Second, we compare BrainLLM with a series of concurrent methods available for open-vocabulary language decoding~(see Supplementary Table~10).
Finally, to validate the proposed framework, we also compare BrainLLM against a standard language model without any brain input~(StdLLM)~(see Method, Supplementary Figs. 4-5) and its variants with different architecture selections~(see Supplementary Information B.3).
The performance of BrainLLM is evaluated from three perspectives:
(1)~win rate: whether BrainLLM has a higher likelihood of generating the perceived continuation than the control model~(PerBrainLLM);
(2)~language similarity metrics~(BLEU, ROUGE, and word error rate~(WER)): measurements of the similarity between the perceived continuation and the generated language;
(3)~human preference: expose the output of BrainLLM and PerBrainLLM to human annotators for judgments on which is semantically closer to the perceived continuation.

\begin{figure*}
    \centering
    \includegraphics[width=\textwidth]{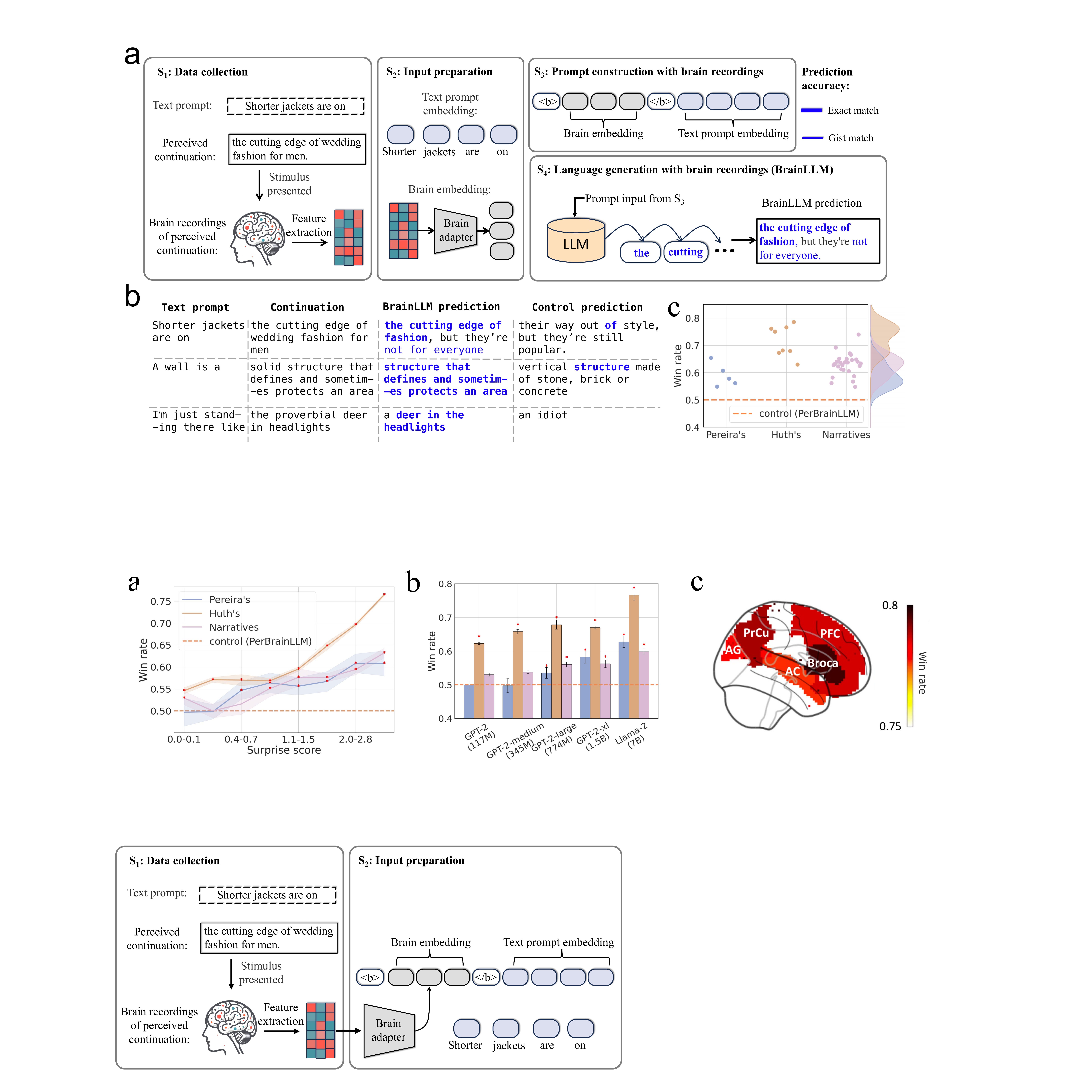}
    \caption{ \textbf{Win rates of BrainLLM vs.\ PerBrainLLM measured by comparing the generation likelihood of the participant's perceived continuation.} Error bars denote mean +/- SEM. The center line, top, and bottom of the box plot represent the group median, 75th percentile, and 25th percentile, respectively. Whiskers are extended to the most extreme data point that is no more than
    1.5 $\times$ interquartile range from the edge of the box. \textbf{(a)} The win rates were significantly higher than 0.5 with $\ac{FDR} < 0.05$~(one-sided non-parametric test) across all datasets and participants.
    Each dot represents the win rate of a single participant in Pereira's dataset~(5 participants), Huth's dataset~(8 participants), and the Narratives dataset~(28 participants).
    \textbf{(b)} 
    The win rate increases as the surprise levels increase.  
    The surprise level quantifies the model's likelihood of generating the continuation stimuli, whereas a higher surprise indicates a greater difficulty in generating the perceived continuation for the LLM.  
    $*$ indicates the win rate is significantly higher than 0.5 with $\ac{FDR} < 0.05$~(one-sided non-parametric test).
    \textbf{(c)} 
    Scatter plot of win rate versus surprise scores for 200 randomly selected tokens. 
    A positive correlation is observed between win rate and surprise, indicating that tokens with higher surprise scores tend to have higher win rates.
    \textbf{(d)}
    The win rate when using brain signals from different cortical regions in a single participant~(participant 1 in Huth's dataset). 
    Brain data~(colored regions) used as input for BrainLLM were partitioned into the Broca’s area, the precuneus~(PrCu), the prefrontal cortex~(PFC), the auditory cortex~(AC), and the angular gyrus~(AG).
    \textbf{(e)}
    The parameter sizes of LLMs exhibit a strong positive correlation with win rates, yielding Pearson's $r$ of 0.886 for Pereira's dataset, 0.953 for Huth's dataset, and 0.923 for the Narratives dataset.
    \textbf{(f)}: The win rate demonstrates a positive correlation with the size of training data. 
    $*$ indicates that the win rate is significantly higher than that of the control.
    For Huth's dataset and the Narratives dataset, which both utilize auditory-based stimuli, the win rate is notably consistent when the datasets are of equivalent size. 
    The total number of data samples within Pereira's dataset, Huth's dataset, and the Narratives dataset amount to 376, 1,039, and an average of 5,546 across participants, respectively.}
    \label{fig:surprise}
\end{figure*}

The averaged win rates of BrainLLM versus PerBrainLLM are 64.9\%, 78.9\%, 66.5\%, on Pereira's dataset, Huth's dataset, and the Narratives dataset, respectively~(Fig.~\ref{fig:surprise}(a)).
This indicates that BrainLLM has a significantly higher likelihood of generating the perceived continuation compared to PerBrainLLM, with the \ac{FDR} $< 0.05$ (one-sided, non-parametric test) on three datasets.
The highest averaged win rate~(78.9\%) is observed on Huth's dataset, which has the largest size of neurological data samples for each participant~(see Fig.~\ref{fig:surprise}(f)).
Similar performance differences have also been observed on language similarity metrics, as shown in Table~\ref{tab:ml}.
This suggests that increasing the size of neurological training data improves the model performance.
Furthermore, we conducted a human evaluation experiment~(detailed in Method) in which 202 annotators recruited from Amazon's Mechanical Turk\footnote{\href{www.mturk.com}{www.mturk.com}} were asked to make a forced-select preference judgment between generation outputs from BrainLLM and PerBrainLLM, or they could opt for ``hard to distinguish'' if no clear preference emerged.
Within the randomly selected sample of 3,000 language pairs generated by BrainLLM and PerBrainLLM from Huth's dataset, the average annotations showed a preference distribution where 48.4\% favored BrainLLM, 39.2\% favored PerBrainLLM, and 12.4\% of the annotators found the pairs indistinguishable. 
The statistical analysis revealed a significant difference in preference between BrainLLM and PerBrainLLM~(p=0.039 using a one-sided non-parametric test).
This human preference between BrainLLM and PerBrainLLM is also found to be associated with higher language similarity metrics~(see Supplementary Information B.4).

Furthermore, we compared BrainLLM with the state-of-the-art method proposed by \citet{tang2023semantic}, which first pre-constructs candidate next tokens with LLM and then adopts a post-hoc selection with brain recordings.
The comparisons are carried out in the aforementioned language generation task and a full-text reconstruction task (as also used in \cite{tang2023semantic}).
In the language generation task, BrainLLM outperforms their approach in all language similarity metrics, with improvements exceeding 40.2\% in BLEU-1 scores~(refer to Supplementary Table~11 and the Supplementary Information B.5).
We further evaluate BrainLLM on a full-text reconstruction task which reconstructs the 10-minute-long story of ``Where There's Smoke'' without any text prompt input~(see Supplementary Information B.6).
We show that BrainLLM can achieve full-text reconstruction by autoregressively treating the generated content as a text prompt for the next step~(as shown in Fig.~\ref{fig:full_text}).
In the full-text reconstruction task, BrainLLM shows comparable performance with \citet{tang2023semantic}'s method but uses a non-classification setup and possesses the ability to access the likelihood of any language segment~(see Table~\ref{tab:ml_full_small}, and Supplementary Table 12).

%% file: meta/3_2_subresults.tex
\subsection*{Language generation performance across continuation with different surprise levels}
LLMs, by predicting the next token with the highest probability, enable the generation of well-structured, coherent language given the text prompt.
This architecture also provides a unified framework for modeling surprise in text continuations by estimating their prediction-error signals~(see Method).
For example, the likelihood of ``meet you'' following ``Nice to'' is higher than ``take chances'', which means that ``meet you" has a lower surprise to LLMs than ``take chances".
Typically, a higher level of surprise indicates that the LLM finds it more ``surprising'' and challenging to generate the perceived continuation. 
We split the test data based on their surprise levels and evaluate BrainLLM on them separately.
As shown in Supplementary Figs. 6-7, both BrainLLM and PerBrainLLM present a performance decrease as the level of surprise increases in terms of BLEU-1.
However, compared to PerBrainLLM, BrainLLM exhibits a more moderate decline in performance.
Furthermore, we examine the win rate of BrainLLM versus PerBrainLLM across perceived continuation with varying levels of surprise, as depicted in Fig.~\ref{fig:surprise}(b).
We observe that the win rate increases as the surprise levels rise.
A significant positive correlation exists between the surprise level and the win rate, with Pearson's $r$ = 0.09, 0.15, and 0.08 in Pereira's, Huth's, and the Narratives datasets, respectively~($\ac{FDR} < 0.05$ in all datasets).  
This suggests that when an LLM deems the perceived continuation as unexpected, the information decoded from brain recordings can significantly enhance the generation process.
Moreover, word tokens exhibiting higher levels of surprise and higher concreteness~\cite{brysbaert2014concreteness} are associated with increased win rates, with Pearson's $r$ of $0.152$ and $0.305$, respectively~(see Fig.~\ref{fig:surprise}(c) and Supplementary Fig. 8).
This suggests the effectiveness of BrainLLM for tokens with more precise meanings.
For instance, concrete nouns such as ``chamber'' and ``leaving'' have higher win rates compared to function words like "the" and "are". 

\subsection*{Effect of text prompt}
Typically, LLMs generate language as a continuation of the given text prompt. 
Existing \ac{NLP} research~\cite{kaplan2020scaling} has shown that the generation accuracy improves when given a longer length of text prompt~\cite{kaplan2020scaling}.
The integration of brain recordings into LLM generation raises a critical question: How does the length of the text prompt affect the performance of BrainLLM? 
Furthermore, how does the BrainLLM perform in scenarios where there is no text prompt provided?
We present the BLEU-1 score of BrainLLM and PerBrainLLM with different lengths of text prompts in Supplementary Figs.~9-10, and the win rate of BrainLLM versus PerBrainLLM is shown in Supplementary Fig.~11.
A negative correlation exists between the length of the text prompt and the win rate, with Pearson's $r$ values of $-0.013$, $-0.059$, and $-0.060$ in Pereira's, Huth's, and the Narratives datasets, respectively.
This observation can be partially explained by the fact that longer text prompts provide LLMs with more contextual information, resulting in a lower level of surprise for the perceived continuation~\cite{ganguli2022predictability, goldstein2022shared}, and consequently reducing the importance of brain input information~(see Supplementary Fig. 12 for the relationship between text length and surprise level).
Additionally, \citet{tikochinski2024incremental} suggest that LLMs can process large contextual windows while the brain may preferentially focus on the content perceived most recently. 
This divergence could also affect the effectiveness of feeding representations decoded from brain signals into LLMs.

Furthermore, we investigate language generation from brain recordings without any text prompt~(see Supplementary Table~13).
On the one hand, we observe that BrainLLM significantly outperforms PerBrainLLM on all language similarity metrics.
The win rate of BrainLLM versus PerBrainLLM~(0.8885 in Pereira's dataset, 0.8816 in Huth's dataset, and 0.6728 in the Narratives dataset) is even higher than that of generation with text prompts.
This enhanced performance of BrainLLM versus PerBrainLLM can be explained by the high surprise levels for perceived continuations when no text prompt is given.
However, the language similarity metrics for the generation without text prompts are lower than those with text prompts.
This indicates that generating language solely based on brain input and without any text prompt is still challenging.

\subsection*{Impact of LLM with different parameter sizes}
We conducted our main experiments based on Llama-2~\cite{touvron2023llama}, which is one of the state-of-the-art LLMs with a large number of parameters, i.e., 7 billion~(7B). 
To study the impact of LLM with different parameter sizes,
we tested a series of generative LLMs constructed with different parameter sizes, including GPT-2~(117M parameters), GPT-2-medium~(345M parameters), GPT-2-large~(774M parameters), GPT-2-xl~(1.5B parameters), and the Llama-2~(7B parameters).
Across PerBrainLLM and BrainLLM, language similarity metrics significantly increase as the number of parameters in the LLM increases~(see Supplementary Table 14). 
This observation aligns with established knowledge: LLMs equipped with more parameters demonstrably excel at language generation~\cite{kaplan2020scaling, zhao2023survey}.
Interestingly, while the performance of PerBrainLLM improves with the increase in the number of parameters, the win rate of BrainLLM over PerBrainLLM also increases~(see Fig.~\ref{fig:surprise}(e)).
This indicates that LLMs with an increasing number of parameters exhibit amplified benefits from brain input.

\begin{figure*}
    \centering
    \includegraphics[width=\textwidth]{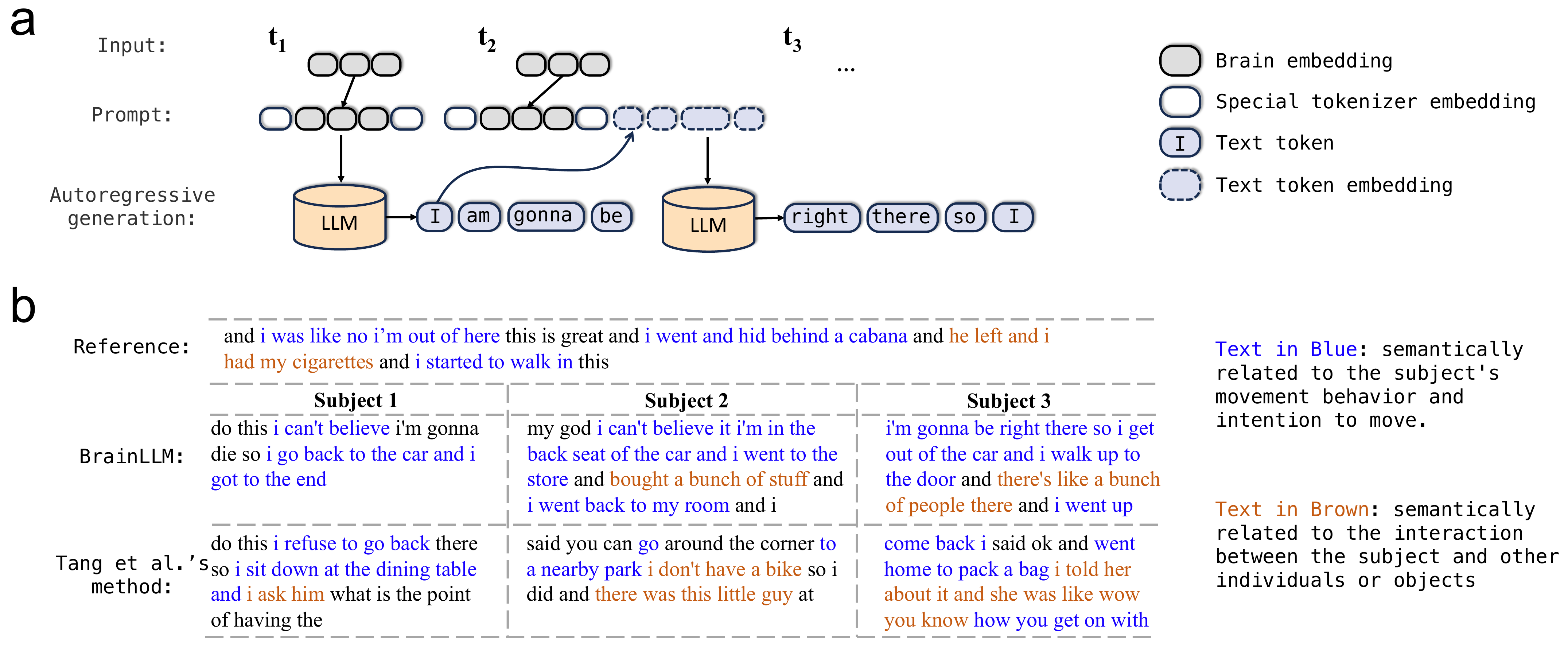}
    \caption{\textbf{Full-text reconstruction with BrainLLM.} \textbf{a,} 
        Illustration of the full-text reconstruction task accomplished with BrainLLM. 
        Each generation step could autoregressively provide the text prompt for the next step.
        \textbf{b,} Examples of full-text reconstruction with BrainLLM and a pre-construction and post-hoc selection method proposed by \citet{tang2023semantic}. 
        Text in \textcolor{blue}{blue} indicates content that is semantically related to the subject's movement behavior and intention to move. 
        Text in \textcolor{brown}{brown} indicates content that is semantically related to the interaction between the subject and other individuals or objects. }
    \label{fig:full_text}
\end{figure*}

\subsection*{Effect of the size of neural activity data for training}
We tested BrainLLM on a variable size of neural activity data and computed its win rate versus PerBrainLLM.
As shown in Fig.~\ref{fig:surprise}(f), the language generation performance steadily increases as the model is trained with more data on Huth's dataset and the Narratives dataset.
Existing studies~\cite{antonello2023scaling, toneva2019interpreting} have found that enlarging the size of neural activity datasets can improve the mapping between language representation in the brain and that in the LLM.
Our results further suggest that expanding the size of neural activity training data also improves language generation performance when jointly modeling the brain representation with LLM. 

\begin{table*}
    \centering
    \caption{
    \textbf{Full-text reconstruction performance for a 10-minute-long story of ``Where There’s Smoke'' in Huth's dataset.} }
    \label{tab:ml_full_small}
    \begin{tabular}{cllll}
    \toprule
    \textbf{Input}  & \textbf{Method}      & \textbf{BLEU-1} & \textbf{WER}    & \textbf{METEOR} \\ \hline
    \multirow{2}{*}{Null}  & Classification~\cite{tang2023semantic} & 0.1908$^{*}$ & 0.9637$^{*}$ & 0.1323$^{*}$ \\                       & BrainLLM    & 0.1417 & 0.9569 & 0.1181 \\ \hline
    \multirow{2}{*}{Subject 1} & Classification~\cite{tang2023semantic} & 0.2331$^{*}$ & 0.9407$^{*}$ & 0.1621$^{*}$ \\  
                               & BrainLLM    & \textbf{0.2539} & \textbf{0.9158} & \textbf{0.2078} \\ \hline
    \multirow{2}{*}{Subject 2} & Classification~\cite{tang2023semantic} & 0.2426$^{*}$ & 0.9354$^{*}$ & 0.1677$^{*}$ \\ 
                               & BrainLLM    & \textbf{0.2518} & \textbf{0.9259} & \textbf{0.2031} \\ \hline
    \multirow{2}{*}{Subject 3} & Classification~\cite{tang2023semantic} & 0.2470  & 0.9243$^{*}$ & 0.1703$^{*}$ \\ 
                               &  BrainLLM    & \textbf{0.2497} & \textbf{0.9190}  & \textbf{0.2180}  \\ \hline
    \end{tabular}
    \begin{tablenotes}
        \item $*$ indicates the performance difference between the pre-construction and post-hoc selection method proposed by \citet{tang2023semantic}~(denoted as ``Classification'') and BrainLLM is significant at $p < 0.05$~(paired t-test).
        A floor for each metric was computed by scoring the mean similarity between the actual stimulus words and a sequence generated from a language model without using any brain data~(``Null''). 
        Here \citet{tang2023semantic} uses a private language model trained on a corpus composed of Reddit stories, which exhibit a similar style with the subject-perceived story content.
        On the other hand, we use a publicly available language model GPT2-xl, which is trained in a general corpus and therefore shows worse performance when compared to ``Classification'' when no brain input is given.
        However, with brain responses collected from human subjects, the proposed BrainLLM shows comparable performance in terms of language similarity metrics with ``Classification''.
    \end{tablenotes}
\end{table*}

\subsection*{Language generation across cortical regions}
We explore how language can be generated with brain recordings collected from different cortical regions as input.
Fig.~\ref{fig:surprise}(d) presents the win rate of BrainLLM versus PerBrainLLM with Broca’s area~\cite{musso2003broca}, the precuneus~(PrCu)~\cite{chee2004left}, the prefrontal cortex~(PFC)~\cite{gabrieli1998role}, the auditory cortex~(AC)~\cite{salmelin1999native}, and the angular gyrus~(AG)~\cite{van2016higher,price2015converging} for a participant~(subject 1) from Huth's dataset.
We observe that BrainLLM significantly outperforms PerBrainLLM in all language processing regions, with its highest score of 0.8012 observed in Broca's area. 
This performance even surpasses the results achieved using responses from all cortical regions.
A partial explanation is the higher information-to-noise rate in these language-related regions and the information loss from dimensionality reduction while using all cortical regions.
Nonetheless, to preclude bias in selecting regions of interest (ROIs), results using responses from all cortical regions are reported in the main findings.
Existing research has shown that during language processing, a substantial portion of the cortex is engaged~\cite{lerner2011topographic,binder2011neurobiology}. 
This suggests that different cortical regions related to language might encode overlapping or similar language representations~\cite{keller2001neural}, potentially facilitating language generation using just a single cortical area. 
These findings have also been observed in prior research on brain language decoding using a pre-construction and post-hoc classification approach~\cite{tang2023semantic, caucheteux2022brains}.

%% file: meta/4_discussion.tex
\section*{Discussion}
\label{sec:discussion}
Our study demonstrates that language can be directly generated with brain recordings as input, rather than through selection from pre-constructed language candidates. 
To accomplish this, we devise an approach that jointly models brain representation and language representation as input for LLMs.
Unlike a standard LLM that generates only the most likely language continuation according to its training data, the generation output of BrainLLM is more aligned with the semantic text content perceived by human participants. 
Using a prompt tuning protocol~\cite{liu2022p,liu2023gpt}, BrainLLM has approximately only 6 million trainable parameters, which is much smaller than Llama-2's 7 billion parameters. 
This parameter size matches existing models like ridge regression commonly used for brain decoding~(e.g., Tang et al.~\cite{tang2023semantic}; Pereira et al.~\cite{pereira2018toward}), yet achieves direct language generation without restricting the generation process on a selection of a pre-defined pool of candidates.

\subsection*{How can we integrate human brain representations into computational language generation models?}
Previous work has shown that the representations in language models and the human brain can be mapped to each other~\cite{toneva2019interpreting,toneva2021bridging,schrimpf2021neural,hale2022neurocomputational,anderson2021deep,sun2020neural}.
Key findings from these studies include exploring how training language models can enhance this mapping~\cite{aw2023training}, and whether brain representations can be used to improve the representation learning in language models~\cite{toneva2019interpreting, goldstein2022shared}.
Our approach differs from the above as the representation alignment between the brain recordings and the language representation in LLMs does not necessarily mean that one can be used to generate the other within a computational framework. 
BrainLLM demonstrates the feasibility of using representations decoded from the brain to enrich the contextual information as input for LLMs,
which is typically based only on text modalities. 
This approach enables LLMs to generate coherent language continuations that match the semantics perceived by human participants~\cite{liu2022p}.

The success of BrainLLM can be attributed to two key factors.
Firstly, the information encoded in the human brain often encompasses contextual and situational semantics~\cite{goldstein2022shared,pereira2018toward}. 
The evidence on the mapping between brain representation and language model representation suggests that contextual and situational semantics can potentially be learned by BrainLLM, enabling effective end-to-end next-token generation training.
Secondly, the increase of language model parameters has given rise to advanced capability in ``few-shot learning'' or ``in-context learning''~\cite{lester2021power}.
BrainLLM uses this capability to backpropagate gradients to train the contextualized representations learned with an fMRI dataset smaller than those typically required for most NLP tasks.
Our experiments also show that language models with increasing model parameter sizes achieve greater performance improvements in BrainLLM than in PerBrainLLM.

\subsection*{Comparison with previous work}
Most existing studies treat the language reconstruction task in a classification setup, which involves pre-defining a set of semantic candidates~(e.g., words~\cite{mitchell2008predicting}, concepts~\cite{pereira2018toward}, sentences~\cite{sun2019towards}) and employing a mapping function to determine which candidate best matches the recorded brain activity.
This setup implies that these methods are incapable of constructing candidates beyond pre-definited sets.
An exception is a recent study~\cite{tang2023semantic} that successfully constructs continuous semantic candidates by first pre-generating several candidate tokens with LLMs, and then selecting from the candidates with brain recordings.

BrainLLM is markedly different from the above studies in that it directly uses the representation decoded from the brain as input to the generative language model.
Such a generative paradigm endows it with the following unique properties:
First, the generative paradigm implies that language reconstruction can be achieved by identifying the correct token without relying on potentially incorrect pre-selected or pre-generated candidates.
The generation process can be considered as selecting the highest probability token from a vocabulary of 32,000 tokens, which exceeds the usual range of 2--50 candidates in previous studies with a classification setup.
At the same time, BrainLLM achieves a top-1 accuracy of up to 65.8\% on the best-performing Pereira's dataset, with accuracy exceeding 40\% across all three datasets~(see Supplementary Fig.~13).
Second, BrainLLM can quantify the generation likelihood of any semantic content rather than a limited number of semantic candidates.
This feature can help neurolinguistic analysis by comparing the generation likelihoods associated with contents with different linguistic characteristics.
Last, existing literature suggests a connection between brain signals and the computation of generative LLMs~\cite{toneva2021bridging,goldstein2022shared}.
The scaling capabilities of BrainLLM in terms of the data size and the parameter size also suggest better adaptability of brain modalities in combination with generative AI models. 

In recent years, many studies in the field of generative AI have inspired and advanced the research in brain decoding.
Generative AI models offer a new pathway for decoding information from the brain, bypassing traditional classification setups.
For example, in addition to the language reconstruction explored in this paper, visual reconstruction from brain data has also progressed from classification-based models~\cite{nishimoto2011reconstructing} to diffusion-based generative models~\cite{scotti2024mindeye2, scotti2024reconstructing, ozcelik2023brain}.
The adoption of generative AI extends beyond information decoding from the brain; it has been shown to elucidate the functional organization of the human visual cortex~\cite{luo2024brain}.
On the other hand, some research has explored why brain recordings have the potential to be jointly modeled with these computational generative models.
For example, \citet{goldstein2022shared, lupyan2015words, clark2013whatever} have shown that the human brain exhibits a tendency to predict the next word, a phenomenon supported by various studies.
Therefore, we believe that the generative reconstruction approach is a promising direction for investigating the perception of information in the brain and could extend beyond the specific model architectures tested here~(See Supplementary Information B.7, and Supplementary Table 15 for a more comprehensive overview).

\subsection*{Implications and future extensions}
Our study illustrates the feasibility of direct language generation from brain recordings and highlights their differences and superiority over previous methods.
Due to the advantages of the generative paradigm, BrainLLM can serve as a superior alternative to traditional classification-based approaches, especially in BCI applications where the user instructions cannot be confined to a pre-defined candidate set.
For example, BrainLLM can help an individual with aphasia to communicate in an open-world environment, without learning a predefined set of user instructions~(see ethics discussion in Supplementary Information B.8).
Despite the superior performance of BrainLLM, open-vocabulary decoding remains highly challenging at a level that could immediately lead to practical applications.
We observe that in the full-text reconstruction task, the output of BrainLLM is still far from perfect matching with the ground truth content~(see Supplementary Table 16).
One promising future direction is to integrate BrainLLM with external modules to infer text prompts and enhance the language generation process, such as incorporating other types of brain-computer interfaces (BCIs).
For example, BCIs based on motor representations~\cite{willett2021high,zhu2010survey,metzger2023high} or attempted language production~\cite{anumanchipalli2019speech} have demonstrated a usable performance, but they require extensive user training and active engagement in the input system, demanding significant user effort~\cite{zhu2010survey,anumanchipalli2019speech}.
In contrast, BrainLLM effectively decodes semantic content from visual and auditory stimuli during participants' perception.
Hence, integrating two types of BCIs could lead to more effective applications: motor-based BCIs generate initial text prompts and enable motor-free language continuation generation, with high-surprise generation steps checked by motor-based BCIs.

Furthermore, BrainLLM essentially quantifies the generation likelihood of participants' perceived continuation when given a text prompt.
Therefore, it can be used to investigate the semantic information encoded in the human brain without a limited set of pre-defined language stimuli.
As the first step, this paper investigates the performance gain brought by brain signals across different surprise levels, context lengths, and different brain regions.
This method can also extend the existing paradigms on studying the representation and perception of language in the brain. 
For example, in neurolinguistic studies~\cite{kutas1980reading}, researchers usually manipulate and pre-define language stimuli with various linguistic characteristics to study their effects on brain responses.
BrainLLM allows us to gather brain data in natural reading settings and analyze it by comparing the generation likelihoods of semantic content with varying linguistic features.
Possible insights may include whether different populations have varying expectations for various language contents and which brain regions are more closely related to specific linguistic aspects. 
Additionally, existing studies have shown that semantic information in the human brain is context-aware~\cite{caucheteux2022brains}, e.g., the brain response to ``flat'' is different in ``flat object'' and ``flat emotion''. 
Since our method is also a context-based~(text prompt) generation, it can be used to explore the impact of contextual information and its effect on brain responses.
An example is exploring the connections between various brain regions and the contextualized semantic aspects by comparing their reconstruction performance.

Last, several studies show that computational language modeling can gain insights from human responses or feedback to language~\cite{ouyang2022training,stiennon2020learning}, especially brain responses~\cite{toneva2021bridging}.
Our experiments show that personalized brain recordings may refine the language generation process, especially when the likelihood of the ground-truth output is low for an LLM. 
This suggests the possibility of training better language models, or at least model with more personalized generation ability that take into account individual variation in brain responses.
For instance, BrainLLM's estimated generation likelihood can facilitate the training of LLMs to produce content that aligns more closely with human expectations.
Training an LLM to align with human expectations has shown its effectiveness with behavioral signals as input and a reinforcement learning technique~\cite{bai2022training}. 
However, while behavioral signals offer only one-dimensional preference feedback, BrainLLM has the potential to provide multi-dimensional feedback across the entire vocabulary distribution, which can be more informative for model training.

%% file: meta/5_method.tex
\section*{Methods}\label{sec:methods} 
We formalize the task of language generation from brain recordings and then detail and justify the different components of BrainLLM, followed by describing the datasets, training, and evaluation.

\subsection*{Task formalization}
Given a text prompt $W$ composed of a sequence of tokens $\{w_1,w_2,w_3,\ldots,w_n\}$, the task objective is to predict its continuation $M = \{m_1,m_2,\ldots,m_k\}$ with the participants' brain recordings while they are perceiving the stimuli constructed with the continuation content $M$.
In this paper, we refer to $M$ as the ``perceived continuation''.
The brain recording $B = \{b_1, \ldots, b_t\} \in \mathbb{R}^{t\times c}$ is a sequence of features extracted from \ac{BOLD} signals, with $c$ being the number of neurological features, and $t$ being the number of time frames in which brain recordings are collected.
We segment $t$ time frames after the stimuli presentation of the perceived continuation.
This segmentation takes into account the delayed effect of \ac{BOLD} signals~\cite{mitchell2008predicting}~($t$ is set to 4, consistent with existing work~\cite{tang2023semantic,toneva2021bridging}). 
The language generation task aims to learn an autoregressive function $F$ that can generate the perceived continuation $M$ one token at a time, utilizing the text prompt $W$ and the brain recording $B$ as inputs. 
This process can be formalized as $\hat{m}_i = F(\{w_1, \ldots, w_n, \hat{m}_1, \ldots, \hat{m}_{i-1}\}, B; \Theta)$, where $\hat{m}_i$ is the $i$-th token generated by the model, and $\Theta$ is the model parameters. 

The language generation ability of BrainLLM is then evaluated in two settings.
The first is to evaluate its performance in predicting the perceived continuation with the ground truth text prompt and the brain input~(i.e., language continuation generation or language generation).
The second is using BrainLLM in an autoregressive manner in which each generation step could autoregressively provide the text prompt for the next step~(full-text reconstruction).
Despite the superior performance of BrainLLM, open-vocabulary decoding with only brain recordings remains highly challenging at a level that could immediately lead to practical applications. 
Therefore, we constructed the above two settings to study the usability of BrainLLM with both machine-based evaluations~(win rates and language similarity metrics) and human evaluations~(see Measurements).

\subsection*{Model}
\subsubsection*{Large language model (LLM)}
In our study, we used the \acp{LLM} released on Huggingface~(\href{https://huggingface.co/models}{https://huggingface.co/models}), namely Llama-2~(\href{https://huggingface.co/meta-llama/Llama-2-7b}{https://huggingface.co/meta-llama/Llama-2-7b}) and the GPT-2 series~(\href{https://huggingface.co/gpt2}{https://huggingface.co/gpt2}).
The GPT-2 series and Llama-2 were selected for our experiment due to their open-source accessibility and extensive utilization in the realm of LLMs.
As of December 2023, they are among the top 10 most downloaded text generation models on Hugging Face.\footnote{\href{https://huggingface.co/models?pipeline_tag=text-generation\&sort=downloads}{https://huggingface.co/models?pipeline\_tag=text-generation\&sort=downloads}}
These \acp{LLM} function in a similar way. 
Typically, they first convert the input tokens into a series of latent vectors with an embedding layer.
Then, these vectors are fed into a multi-layer neural network that uses multi-head self-attention to aggregate the representations of each vector in a sequence~\cite{vaswani2017attention}.
Based on this architecture, for any input sequence of tokens $S=\{s_1, s_2, \ldots, s_n\}$ with length $n$, the LLM can estimate a prior probability distribution $P(s_{n+1}\mid S)$ for the next token $s_{n+1}$ over the given sequence $S$. 
This probability estimation function $P$ serves as a mechanism for autoregressive language generation. 
Conventionally, the input tokens $S$ are text-based. 
However, in our approach the brain recordings are incorporated into the construction of sequence $S$, enabling language generation that is aware of the brain input. 
Additional details regarding the statistics, and abilities of different LLMs are provided in Supplementary Information B.9 and Supplementary Table~17.

\subsubsection*{Input preparation}
First, the text prompt is directly fed to the LLM's embedding layer $f_w$ to transform the tokens into latent vectors $V^{W}=\{v^W_1, \ldots, v^W_n\} \in \mathbb{R}^{n\times d}$, where $n$ is the number of tokens, and $d$ is the embedding size.
Second, a brain adapter $f_b$ is devised to embed the brain recording into the same latent space with the dimension $d$.
Specifically, for each $b_i \in B$, the decoder embeds it into the space $\mathbb{R}^d$, which can be formulated as $v^B_i=f_b(b_i)$.
Last, the brain embedding $V^B$ and the text embedding $V^W$ are concatenated together, allowing the LLM to perceive modalities from the brain and the text in a unified representation.
To differentiate between the two modalities effectively, we introduce two special tokens, i.e., $\langle brain\rangle$ and $\langle/brain\rangle$, to indicate the beginning and end of the brain embedding. 
The special tokens are randomly initialized as one-dimensional vectors $v^{\langle brain\rangle}$ and $v^{\langle /brain\rangle}$, respectively. 
These vectors have the same number of dimensions $d$ as the token embeddings in LLM.
As a result, the input sequence $I$ can be formulated as $I=\{v^{\langle brain\rangle},v^B_1,\ldots,v^B_t,v^{\langle/brain\rangle},v^W_1, \ldots, v^W_n\}$.

\subsubsection*{Brain adapter}
The brain adapter is a deep neural network $f_b$, with the brain recording $B=\{b_1,\ldots,b_t\} \in \mathbb{R}^{t\times c} $ as input and the brain embedding $V^{B}=\{v^B_1,\ldots,v^B_t\} \in \mathbb{R}^{t\times d}$ as output, where $d$ is the LLM's embedding size.
The architecture of the brain adapter is chosen from a range of candidates~(see Supplementary Information B.3, Supplementary Fig. 14, and Supplementary Table 18). 
Unlike LLMs that connect with other modalities~\cite{duan2024dewave,fathullah2024audiochatllama,chu2024qwen2,huang2024language}, the brain adapter in BrainLLM models brain representations non-linearly, taking into account the delay effects of BOLD signals and adopted position embedding for sequence modeling.
Specifically, $f_b$ comprises (1)~a position embedding $P=\{p_1, \ldots, p_t\} \in \mathbb{R}^{t \times c}$ that captures and represents the chronological order during the collection of \ac{BOLD} signals, and (2)~a multi-layer perceptron network $f_m$ designed to transform the brain representation into the latent space that is shared with the text modalities. 
The position embedding is initialized using a uniform distribution and set to be trainable. 
Element-wise addition is applied where each position embedding $p_i \in P$ is added to its corresponding \ac{BOLD} features $b_i \in B$.
The multi-layer perceptron network $f_m$ is constructed with an input layer and two hidden layers that have the same dimension $c$ as the input \ac{fMRI} features, as well as the output layer with the dimension of $d$.
A ReLU~\cite{fukushima1980neocognitron} is used as the activation function.
Formally, the \ac{BOLD} features corresponding to the $i$-th time frame, denoted as $b_i$, is input into the brain adapter $f_b$, which can be expressed as $v^B_i=f_b(b_i)=f_{m}(p_i+b_i)$.
The output vector embedding $v^B_i$, with its dimension tailored to the LLM's embedding size, can be further adopted to construct the input with the text modalities.

\subsubsection*{Training objective}
Inspired by the prompt tuning technique~\cite{liu2023pre}, the training of our proposed model involves a warm-up step, followed by a main training step.
The warm-up step aims to align the distribution of the brain embedding with that of the text token's embeddings, ensuring that the brain embedding is primed for integration with the text prompt embedding.
This step aims to develop an adapter that extracts information from brain signals relevant to the current semantic context, thereby enhancing the robustness of modeling noisy fMRI signals. 
To streamline the process and enable training without leaking information about the perceived continuation, each $v^B_i \in V^B$ is simply mapped to the mean value of the corresponding text prompt embeddings, i.e., $\frac{1}{n}\sum_{j=1}^n v^W_j$. 
The \ac{MSE} loss is used during the training process of the warm-up step:
\begin{equation}
L_\mathit{MSE} = \frac{1}{t} \sum_{i=1}^{t} (v^B_i - \frac{1}{n}\sum_{j=1}^{n}v^W_j )^2
\end{equation}
Then, we construct the input sequence $I$ combined with both brain and text modalities.
The LLM utilizes a transformer architecture for autoregressive generation based on the input sequence $I$. 
The main training target is selected as maximizing the generation likelihood of the perceived continuation:
\begin{equation}
\mathop{max}_{\Theta} \sum_{i=1,2, \ldots,k} \log(P(m_i\mid I, \{ m_1, \ldots, m_{i-1} \};\Theta))
\end{equation}
where $\Theta=\{\Theta^{LLM}, \Theta^{f_b}, \Theta^{sp}\}$ is the model parameters, $\Theta^{LLM}$, $\Theta^{f_b}$, and $\Theta^{sp}$ are the parameters of the LLM, the brain adapter, and the special tokens $\langle brain\rangle$ and $\langle/brain\rangle$, respectively.
During the main step, we retain the inherent knowledge of the LLM while learning useful information from a limited number of data samples with the ``prompt tuning'' technique~\cite{liu2023gpt}. 
This technique involves keeping the parameters of the LLM unchanged, and instead, fine-tuning only the input representation, i.e., $\Theta^{f_b}$, and $\Theta^{sp}$ in our task. 
By doing so, the brain adapter learns to decode information from the human brain recordings to guide the LLM in generating outputs that closely resemble the perceived continuation. 
This technique has been experimentally validated to be more effective than fine-tuning all LLM parameters~(see Supplementary Information B.3 and Supplementary Table 19).

\subsection*{Datasets \& preprocessing}
We test BrainLLM on three public fMRI datasets, Pereira's dataset~\cite{pereira2018toward}, Huth's dataset~\cite{lebel2023natural}, and the Narratives dataset~\cite{nastase2021narratives}.
All datasets, along with their associated studies, received approval from ethics committees and are accessible for basic research. 
Informed consent was secured from every human research participant.
Pereira's dataset collects participants' BOLD signals while viewing visual stimuli composed of Wikipedia-style sentences.
Consistent with previous work~\cite{luo2022cogtaskonomy}, the brain data of participants who both participated in experiments 2 and 3 were selected in this paper.
This involves 5 participants, each responding to 627 sentences. 
The released beta coefficient brain images~(see the original paper~\cite{pereira2018toward}) corresponding to each sentence are used in our study.
Huth's dataset and the Narratives dataset contain BOLD responses recorded while participants listened to auditory language stimuli of narrative stories.
The officially released preprocessed motion-corrected version of these datasets is adopted in our study~(\href{https://openneuro.org/datasets/ds003020/}{https://openneuro.org/datasets/ds003020/} and \href{https://openneuro.org/datasets/ds002345/}{https://openneuro.org/datasets/ds002345/}).
Huth's dataset includes data from 8 participants, each listening to 27 stories. 
Consequently, each participant contributed 6 hours of neural data, amounting to a total of 9,244 \acp{TR}.
The Narratives dataset initially included 365 participants, from which we selected 28 individuals who engaged in at least three story stimuli.
Among them, eight participants took part in 4 stories, while 20 participants took part in 3 stories, with an average of 1,733 TRs collected from each participant. 
Additional details regarding the statistics, approvals, pre-processing, and language stimuli for these datasets are provided in the Supplementary Information A and Supplementary Table 20.

To efficiently manage and analyze the fMRI data, we consistently apply dimension reduction to $c=1,000$ dimensions across all datasets for the whole-brain BOLD features.
The dimension reduction is obtained by applying principal component analysis~\cite{abdi2010principal} to the preprocessed \ac{BOLD} features.
When conducting analysis on a single brain region, the original signal was directly used without dimension reduction.
Consequently, we constructed the data samples for the language generation task with the \ac{BOLD} features in each time frame, corresponding stimuli presented to the participant~(perceived continuation), and the text prompt~(if any) that preceded the stimuli.
Pereira's dataset consists of participants' brain recordings of individual sentences, each presented without overlap.
We split each sentence into three pieces with approximately equal number of tokens. 
Two unique data samples are constructed by treating the first third as the text prompt and the second third as the perceived continuation, as well as combining the first two-thirds as the text prompt and using the last third as the perceived continuation. 
For Huth's dataset and the Narratives dataset, the language stimuli were presented to the participants continuously.
Therefore, we split the dataset by treating each \ac{TR}~(2s in Huth's dataset and 1.5s in the Narratives dataset) as a time frame.
The perceived content during each time frame is selected as a perceived continuation. 
Then we used a sliding window ranging from 1 to 3 TRs to select the language stimuli preceding the appearance of the perceived content as the text prompt.
This step created 3 data samples for each time frame.
The construction of data samples aims to create as many samples as possible with limited neurological data and ensure that the model is adept at handling text prompts of varying lengths. 
After that, the data samples are split into training, validation, and testing sets with a size roughly proportional to 3:1:1, respectively.
The splitting ensured that there was no overlap of perceived continuation and brain recordings among the training, testing, and validation sets.
Additional details and examples for the dataset construction are provided in Supplementary Information B.1.

\subsection*{Training and inference protocols}
We trained BrainLLM with the Adam optimizer~\cite{kingma2014adam} using a learning rate of $1\times 10^{-4}$ and a batch size of 8. 
The batch size is set to 8 as the significant graphics memory demands of the LLM preclude the use of a bigger batch size.
The training of the warm-up step was stopped after ten epochs.
The training of the main step was stopped when no improvement was observed on the validation set for ten epochs, while the test set was never used during the training process. 
The entire training process was conducted on 16 A100 graphics processing units with 40 GB of memory and took approximately 14 hours to complete.

For inference on the test set, we adopted a beam search method.
We maintain a beam containing the five most likely sequences and generate a continuation for each beam at each generation step. 
Then we truncate the number of tokens under the given TR for evaluation.
This truncation remains consistent across BrainLLM, its control, and the re-implementation of baselines.
In the full-text reconstruction task, we use a word rate model following existing research~\cite{tang2023semantic} to predict the number of tokens perceived at each TR, and generate an equivalent number of tokens at each step.
Discussions on the hyper-parameter selection are provided in Supplementary Information B.10 and Supplementary Table 21.

\subsection*{Full-text reconstruction}
We investigated the application of BrainLLM in the reconstruction of full-text content.
Initially, assume the brain recordings corresponding to the first time frame are $\{b_{0,1}, ..., b_{0,t}\}$, where $t=4$ is the segmentation time window when taking into account the delayed effect of BOLD signals.
We adopt a word rate model $\text{WR}$ following existing work~\cite{tang2023semantic}, which predicts the length $l_0$ of word tokens perceived by an individual within a given time frame using brain recordings as input:
\begin{equation}
    l_0 = \text{WR}(\{b_{0,1}, ..., b_{0,t}\})
\end{equation}
Subsequently, based on the prompt input decoded from the brain recordings, i.e., $\{ v^{\langle brain\rangle}, v_{0,1}^B, \ldots , v_{0,t}^B, v^{\langle /brain\rangle}\}$ and the predicted word rate $l_0$, we generate $l_0$ tokens with the LLM at the first time step:
\begin{align}
    M_0 = \text{LLM}(\{ v^{\langle brain\rangle}, v_{0,1}^B, \ldots , v_{0,t}^B, v^{\langle /brain\rangle}\}), \nonumber &
    \\i \in \{1,2,\ldots,m_0\}&
\end{align}
where $M_0=\{m_{0,1},...,m_{0,l_0}\}$ is the $l_0$ tokens generated with the prompt input.
Following this, at the $k^{th}$ time frame, continuations are produced based on the brain recordings at the $k^{th}$ frame $\{b_{k,1}, \ldots, b_{k,t}\}$ and the tokens generated in the previous time steps $\{w^k_{1}, \ldots, w^k_{s}\}$, where $s$ is the window size to truncate the previously generated tokens. 
At the $k^{th}$ time step, the input for the LLM comprises the embeddings of these tokens and the brain input:
\begin{align}
    M_k = \text{LLM}(\{ v^{\langle brain\rangle}, v_{k,1}^B, \ldots , v_{k,t}^B, v^{\langle /brain\rangle}\nonumber &
    \\, v^{w^k}_1,\ldots, v^{w^k}_s\}), i \in \{1, 2, \ldots, m_k\}&
\end{align}
where $M_k=\{m_{k,1},...,m_{k,l_k}\}$ is the tokens generated in the $k^{th}$ time step, $l_k$ is the predicted word rate in the $k^{th}$ time frame, $\{v^{w^k}_1,\ldots, v^{w^k}_s\}$ are the word embeddings of the previously generated tokens.

The newly introduced hyperparameters for the full-text reconstruction involve the size of the time window and the beam size when conducting beam search for content generation~\cite{zhao2023survey}. 
We tested the size of the time window from $\{5, 10, 20\}$ and the beam size from $\{3, 5, 10\}$. 
Ultimately, the selected optimal hyperparameters are a time window of 10 and a beam size of 3.

\subsection*{Machine evaluation}
We investigate BrainLLM and its baselines and controls based on two machine evaluation measurements, i.e., (1)~surprise and win rate, and (2)~language similarity metrics.

The surprise and win rate are measured based on the likelihood of BrainLLM generating the perceived continuation.
Given a sequence of tokens, LLM induces a distribution of probabilities for all possible following continuations.
The likelihood of a possible continuation is the multiplicative product of the probabilities of generating each token in the continuation.
Typically, the negative logarithmic cross-entropy likelihood of the perceived continuation in this distribution is adopted as the surprise measurement~\cite{meister2021language}:
\begin{equation}
\mathit{surprise} = -{\sum_{i=1,2,\ldots,k} \log(P(m_{i}\mid I, \{m_{1},\ldots,m_{i-1})\})}
\end{equation}
where $\{m_{1},\ldots,m_{k}\}$ is the continuation of input sequence $I$.
The higher surprise indicates the language model deems the continuation as more unexpected.
Based on this definition, a more effective language generation model should deem the perceived continuation less surprising.
Consequently, to assess the relative performance of the proposed BrainLLM and its control models, PerBrainLLM and StdLLM, we compare their surprise scores for each perceived continuation within the constructed data sample.
This evaluation metric is known as win rate and has been utilized for performance comparison in brain decoding and encoding research~\cite{mitchell2008predicting,pereira2018toward}.
In addition, we also utilize PerBrainLLM's surprise measurement to examine the impact of surprise on language generation performance, as this measurement represents the language model's surprise for the perceived continuation when brain recordings corresponding to the perceived continuation are not obtained.

The language similarity metrics used in our study are BLEU~(Bilingual evaluation understudy)~\cite{papineni2002bleu}, ROUGE~(Recall-Oriented Understudy for Gisting Evaluation)~\cite{chauhan2022comprehensive}, WER~(Word Error Rate)~\cite{klakow2002testing}, and METEOR~(Metric for Evaluation of Translation with Explicit Ordering)~\cite{chauhan2022comprehensive}.
To avoid potential bias introduced by relying on language representations from LLMs, we refrain from employing metrics such as BertScore~\cite{zhang2019bertscore}, which utilize LLM-derived representations.
BLEU is a metric for measuring the similarity between two text sequences, and is based on the n-gram precision between the generated sequence and reference sequence.
The BLEU score is computed as by:
\begin{equation}
\text{BLEU} = \frac{\text{BP}}{(\text{BP} + (1 - \text{BP}) * (1 - e^{-\ln(r_n) / \ln(m)}))}
\end{equation}
where $r_n$ is the n-gram precision, which is the number of n-grams that match between the generated sequence and the reference sequence,
$m$ is the number of possible n-grams in the reference sequence,
$\text{BP}$ is the brevity penalty, which is a measure of how much shorter the generated sequence is than the reference sequence, which can be measured by:
\begin{equation}
\text{BP} =
\begin{cases}
1 & \text{if } r < c \\
e^{1-r/c} & \text{if } r \ge c
\end{cases}
\end{equation}
We used the unigram variant BLEU-1 in our paper.
WER is calculated as the number of words that are incorrectly recognized divided by the total number of words in the reference sequence, which is measured by:
\begin{equation}
\text{WER} = (\text{substitutions} + \text{deletions} + \text{insertions}) / m
\end{equation}
where $m$ is the number of possible n-grams in the reference sequence, \text{substitutions}, \text{deletions}, and \text{insertions} are the number of substitutions, deletions, and insertions while transforming the generated sequence to the reference sequence.
ROUGE~(Recall-Oriented Understudy for Gisting Evaluation) is another metric for measuring the similarity between two text sequences. 
It is based on the recall of the n-grams in the generated sequence:
\begin{equation}
\text{ROUGE-N }= \frac{r_n}{m}
\end{equation}
where $r_n$ is the n-gram recall, which is the number of n-grams that match between the generated sequence and the reference sequence divided by the total number of n-grams in the reference sequence,
$m$ is the number of possible n-grams in the reference sequence.
We use the unigram variant and the longest common subsequence variant of ROUGE.
The longest common subsequence variant of ROUGE is computed as by:
\begin{equation}
\text{ROUGE-L} = \frac{\text{RLCS}}{m}
\end{equation}
where $\text{RLCS}$ is the length of the longest common subsequence between the generated sequence and the reference sequence.
METEOR is a metric not only considers the exact match of n-grams but also accounts for the proper ordering of them.
METEOR first calculates a parametrized harmonic mean $F_{mean}$ of unigram precision and unigram recall.
Then, the sequence of matched unigrams is divided into the fewest possible number of ``chunks'' to calculate a fragmentation fraction as a penalty.
Finally, the METEOR score is computed as:
\begin{equation}
    \text{METEOR} = (1-\text{penalty})\cdot F_{\text{mean}}
\end{equation}
Since the chunks are few in the language generation task, which renders METEOR meaningless, we only use METEOR in the full-text reconstruction task.
In general, higher scores in BLEU, ROUGE, and METEOR, coupled with a lower score in WER, indicate higher language similarity.
Discussions and extended analysison the measurements are provided in Supplementary Information B.4, Supplementary Table 12, and Supplementary Tables 22-23.

\subsection*{Human evaluation}
202 participants were recruited from Amazon’s Mechanical Turk\footnote{\href{https://www.mturk.com/}{https://www.mturk.com/}} for the human evaluation.
All participants have stipulations of U.S. residents~(based on ownership of a U.S. bank account).
These participants were required to have maintained at least a 90\% approval rate on their previous HITs and to have had a minimum of 1,000 HITs approved historically.
Informed consent was obtained from all participants included in the study. 
This study adheres to the ethical procedures which is approved by the ethics committee of the School of Psychology at Tsinghua University with the identifier 2021 Ethics Approval No. 18. 
All ethical regulations relevant to human research participants were followed.

The human evaluation task is selected as a preference judgment between generation output from BrainLLM and PerBrainLLM.
PerBrainLLM is selected as the control of BrainLLM in the human evaluation study, as their comparison directly demonstrates the impact of utilizing brain recordings corresponding to the perceived continuation.
We randomly sampled 3,000 pairs of generation output from BrainLLM and PerBrainLLM in Huth's dataset for the task.
To mitigate the order effect, each pair of language contents generated from BrainLLM and PerBrainLLM are randomly assigned as ``Text1'' and ``Text2''.
As shown in Supplementary Fig. 15, participants are required to judge which one in a pair~(``Text1'' and ``Text2'') is semantically closer to the perceived continuation~(namely ``Base Text'').
This preference judgment is made by selecting from ``Text1 is better'' and ``Text2 is better'', or the participant can select ``hard to distinguish'' if they find it difficult to judge or deem ``Text1'' and ``Text2'' as equally good.
On average, the participants were paid \$1.0 for each 15 minutes they spent. 
This rate of pay~(\$4.0 per hour) is above the median hourly wage for MTurk HITs. 
All results are included in our analyses.
A one-tailed t-test is implemented to statistically assess the disparity in the preference counts for BrainLLM and PerBrainLLM. 
In this analysis, instances categorized as ``hard to distinguish'' are assigned a midpoint value, equidistant between the two options. 
This approach recognizes the option of ``hard to distinguish''  as representing a balanced or neutral preference. 

\subsection*{LLM control selection}
\label{Standard LLM control}
Instead of using permutated inputs as a control~(PerBrainLLM), utilizing the outputs of a standard LLM~(StdLLM) as a baseline for comparative analysis is a more prevalent practice~\cite{tang2023semantic,xi2023unicorn}.
However, we doubt that this prevalent selection of StdLLM might not be a fair baseline.
We test the performance of PerBrainLLM and StdLLM, finding that PerBrainLLM significantly outperforms StdLLM~(see Supplementary Fig. 16, Supplementary Table 24). 
Notably, a similar phenomenon is observed in the previously proposed method with a pre-construction setup~\cite{tang2023semantic} in our experiment~(see Supplementary Table~25). 
The enhanced performance of PerBrainLLM over StdLLM lies in its ability to generate content that aligns with the common data distribution of language usage in the dataset.
Although PerBrainLLM uses brain recordings that are not aligned with stimuli perceived by an individual for a particular continuation, these contents share similar language usage patterns~(e.g., all stimuli in Pereira's dataset are Wikipedia-style).
We analyze this problem theoretically from a probability perspective and provide more experimental details in Supplementary Information B.2.

\subsection*{Statistics and Reproducibility}
All statistical analyses were performed using the Python~(version 3.8.12) and the packages SciPy~(version 1.9.1), and Statsmodels~(version 0.13.2). 
All bar graphs represent the mean value and the standard error as errors bars, points represent averaged values from an indivual particpants~(e.g., Fig.~\ref{fig:surprise}(a)) or an decoded token target~(e.g., Fig.~\ref{fig:surprise}(c)). 
All statistical analyses for win rates and human evaluations are one-sided tests, while analyses for language similarity metrics are paired tests.
When the data follows a normal distribution, we use the t-test; otherwise, the non-parametric Wilcoxon test is used.
FDR was conducted across the three datasets, with a threshold of 0.05 considered significant.
We also include a structural equation modeling for analyzing Relationship bwteeen the length of the text prompt and the win rate in Supplementary Information B.4 and a analysis of the relationship between different measurements in Supplementary Fig. 17.
Reproducibility was maintained by open-sourced code and data.

\subsection*{Data Availability}
The data for analyses and generating figures are available at: \href{https://doi.org/10.6084/m9.figshare.28352153}{https://doi.org/10.6084/m9.figshare.28352153}~\cite{Ye2025}.
The data from Pereira et al.~\cite{pereira2018toward} is available under the CC BY 4.0 license at the OSF platform~(\href{https://osf.io/crwz7}{https://osf.io/crwz7})~\cite{pereira2018toward_dataset}.  
Huth's dataset~\cite{lebel2023natural} is provided~(in part) by the University of Texas at Austin with a ``CC0'' license at the OpenNeuro platform~((\href{https://openneuro.org/datasets/ds003020/}{https://openneuro.org/datasets/ds003020/})~\cite{lebel2023natural_dataset}. 
The Narratives dataset~\cite{nastase2021narratives} is available under the same universal license at the OpenNeuro platform~(\href{https://openneuro.org/datasets/ds002345/}{https://openneuro.org/datasets/ds002345/})~\cite{nastase2021narratives_dataset}. 
All audio or visual files were provided by the authors of each dataset.

\subsection*{Code Availability}
The code for our paper can be found at \href{https://zenodo.org/records/14838723}{https://zenodo.org/records/14838723}~\cite{my_dataset}.
All code and materials used in the analysis are available under the CC-NC-BY 4.0 license.